\documentclass{article}
\usepackage{spconf,amsmath,graphicx}
\usepackage{caption}
\usepackage{subcaption}

\usepackage{microtype}
\usepackage[T1]{fontenc}
\usepackage{placeins}
\usepackage{latexsym}
\usepackage{graphicx}
\graphicspath{{figures/}}

\usepackage{subcaption}
\usepackage{booktabs}
\usepackage{makecell}
\usepackage{float}
\graphicspath{{figures/}}

\usepackage{mathtools}
\usepackage{amssymb} 
\usepackage{mathrsfs} 
\usepackage{bbm}  
\usepackage{algorithm}
\usepackage{algpseudocode}

\algnewcommand{\IIf}[1]{\State\algorithmicif\ #1\ \algorithmicthen}
\algnewcommand{\EndIIf}{\unskip\ \algorithmicend\ \algorithmicif}

\algnewcommand{\LineComment}[1]{\State \(\triangleright\) #1}
\algnewcommand{\IfThenElse}[3]{
  \State \algorithmicif\ #1\ \algorithmicthen\ #2\ \algorithmicelse\ #3}
  
 \algnewcommand{\IfThen}[2]{
  \State \algorithmicif\ #1\ \algorithmicthen\ #2}

\usepackage{nameref}  

\usepackage{xspace}  
\usepackage{enumitem}

\newcommand{\x}{\boldsymbol x}

\newcommand{\ktr}{{k_\text{train}}}

\makeatletter
\DeclareRobustCommand\onedot{\futurelet\@let@token\@onedot}
\def\@onedot{\ifx\@let@token.\else.\null\fi\xspace}

\makeatother

\usepackage{tikz, pgfplots}
\usepackage{pgffor, pgfmath}

\pgfplotsset{
    compat=1.15,
    grid style={darkgray},
    minor grid style={gray!20},
    major grid style={gray!20},
    axis line style = { darkgray }, 
    every axis plot/.append style={line width=1.5pt, mark options=solid, mark size=4pt},
    legend style={draw = darkgray, rounded corners=0pt, fill = white, font=\Large},
    tick style ={color = gray!30 },
    tick label style={font=\normalsize},
    label style={font=\normalsize},
}

\def\x{{\mathbf x}}

\title{AN EMPIRICAL STUDY OF END-TO-END SIMULTANEOUS SPEECH TRANSLATION DECODING STRATEGIES}
\name{Ha Nguyen$^{1,2}$, 
    Yannick Est{\`e}ve$^2$, 
    Laurent Besacier$^{1,3}$}

\address{
   $^1$LIG - Universit{\'e} Grenoble Alpes, France  \\
    $^2$LIA - Avignon Universit{\'e}, France \\
    $^3$Naver Labs Europe, France}

\begin{document}
%
\maketitle
\begin{abstract}
This paper proposes a decoding strategy for end-to-end simultaneous speech translation. We leverage  end-to-end models trained in offline mode and conduct an empirical study for two language pairs (English-to-German and English-to-Portuguese). We also investigate different output token granularities including characters and Byte Pair Encoding (BPE) units. The results show that the proposed decoding approach allows to control BLEU/Average Lagging trade-off along different latency regimes. Our best decoding settings  achieve comparable results with a strong cascade model evaluated on the simultaneous translation track of IWSLT 2020 shared task.
\end{abstract}
\begin{keywords}
Simultaneous speech translation, end-to-end models, low-latency decoding.
\end{keywords}
\section{Introduction}
\label{sec:intro}

Simultaneous (online) machine translation  consists in generating an output hypothesis before the entire input sequence is available \cite{bangalore2012real, sridhar2013segmentation}. 
To deal with this low latency constraint, several strategies were proposed for neural machine translation with input text~\cite{Ma19acl, Arivazhagan19acl, Ma20iclr}. 
Only a few works investigated low latency neural speech translation~\cite{DBLP:journals/corr/abs-1808-00491,elbayad:hal-02895893,han-etal-2020-end}.
At IWSLT 2020 workshop, a \textit{simultaneous translation track} was proposed and attempted  to stimulate research on this challenging task \cite{iwslt:2020}.
In 2020, the best system of this track \cite{elbayad:hal-02895893} was made up with a cascade of an ASR system trained using Kaldi~\cite{Povey11thekaldi} and an online MT system with \emph{wait-$k$} policies~\cite{Dalvi18naacl, Ma19acl}. Only one end-to-end simultaneous system was proposed \cite{han-etal-2020-end} but its performance was not at the level of the cascaded model.

In this paper, we propose a simple but efficient decoding approach that allows to leverage any pre-trained end-to-end speech translation (offline) model for simultaneous speech translation. We conduct an empirical study of the decoding parameters for different language pairs and for different output token granularities (characters or BPEs). Our contributions are the following:
\vspace{-5pt}
\begin{itemize}
    \item We adapt the algorithm from \cite{han-etal-2020-end} but introduce the possibility to write several output tokens at a time.
    \vspace{-5pt}
    \item We show that it allows to control AL/BLEU trade-off along different latency regimes with a pre-trained end-to-end AST model that does not need to be re-trained in simultaneous mode.
    \vspace{-5pt}
    \item We conduct an empirical evaluation of our decoding strategies for 2 different language pairs (English-to-Portuguese (EN-PT) and English-to-German (EN-DE)) and with different output granularities (characters or BPEs).
    \vspace{-5pt}
    \item We show that our best system, evaluated on the same IWSLT 2020 simultaneous shared task dataset, is competitive with the winning (cascade) system of the IWSLT 2020 shared task \cite{elbayad:hal-02895893}.
\end{itemize}

\section{END-TO-END SIMULTANEOUS DECODING STRATEGIES FOR AST}
\label{sec:simultaneous}

\subsection{Architecture of the end-to-end model}
As mentioned in \cite{nguyen2019ontrac}, we use an attention-based encoder-decoder architecture, whose encoder is a stack of 2 VGG-like~\cite{simonyan2014very} CNN blocks, and then 5 layers of 1024-dimensional BLSTM \cite{LSTM}. Each VGG block consists of two 2D-convolution layers, followed by a 2D-maxpooling layer. These two VGG blocks transform the shape of input speech features from $(T \times D)$ to $(T/4 \times D/4)$, with $T$ being the length of the input sequence (number of frames), and $D$ being the features' dimension.
We use Bahdanau's attention mechanism~\cite{bahdanau2014neural} throughout all the experiments presented in this paper. The decoder is a stack of two 1024-dimensional LSTM layers.

\vspace{-5pt}
\subsection{Simultaneous decoding strategies}

Our end-to-end simultaneous decoding strategy is inspired by \textit{wait-k} decoding strategy introduced for text NMT in \cite{Ma19acl}. It is also built on \cite{han-etal-2020-end} who proposed the only end-to-end approach  at the simultaneous translation track of IWSLT 2020. However,  their models were based on Transformer and a meta-learning approach was needed to obtain decent results in low latency regimes. Transformer-based \textit{wait-k} models  also required re-training in low latency mode in \cite{elbayad:hal-02962195}. In this work, we hypothesize that simpler LSTM encoders might be more robust to limited source context when no re-training is performed in online mode.  Consequently, our (LSTM-based) end-to-end models  trained in offline mode are re-used without any adaptation nor re-training in this work.

For online decoding, we propose a deterministic  strategy\footnote{We leave dynamic decoding with adaptive read/write policy for future work} with the following parameters (see figure \ref{fig:waitk}):

\begin{figure*}
    \captionsetup[subfigure]{justification=centering}
    \centering
    \begin{subfigure}[b]{0.45\textwidth}
        \includegraphics[width=\textwidth]{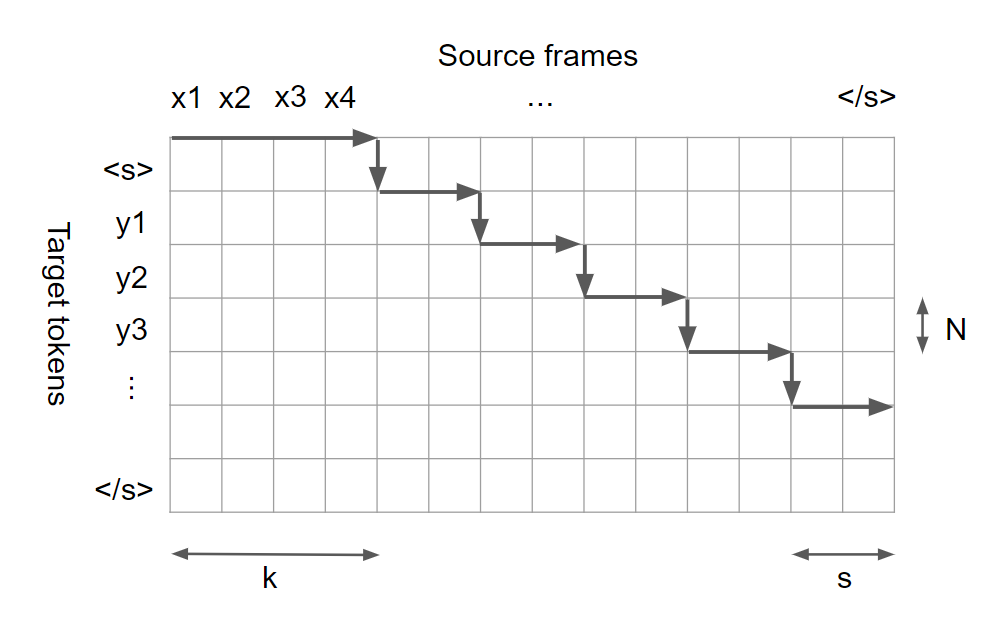}
        \caption{N=1}
        \label{fig:waitkN1}
    \end{subfigure}
    \begin{subfigure}[b]{0.45\textwidth}
        \includegraphics[width=\textwidth]{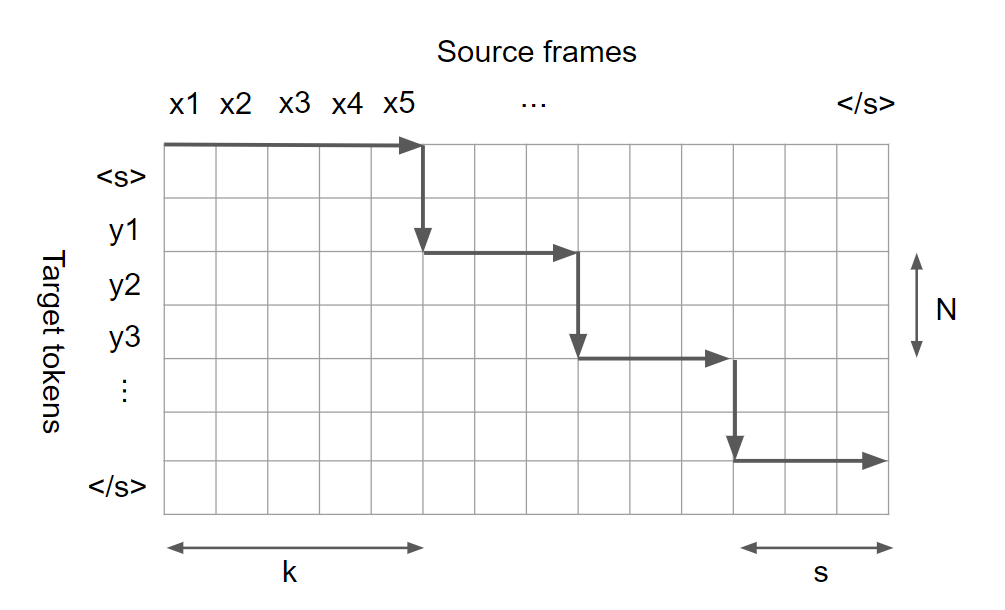}
        \caption{N=2}
        \label{fig:waitkN2}
    \end{subfigure}
    \centering
    \caption{Illustration of our simultaneous decoding strategy: $k$ source frames are read before 1st decoding step where $N$ tokens are generated; then $s$ additional frames are read before the next decoding step, etc. Different values of $(k, s, N)$ are displayed in (a) and (b). }
    \label{fig:waitk}
\end{figure*}

\vspace{-10pt}

\begin{itemize}
    \item $k$ (\textit{wait} parameter) denotes the number of acoustic frames (at the beginning of the input speech features sequence) read before writing the first output token (k=100 or 200 frames in our experiments which is equivalent to 1s or 2s),
    \vspace{-5pt}
    \item $s$ (\textit{stride} parameter) represents the number  of acoustic frames in the input speech features sequence to be consumed in order to produce each new target token (s=10 or 20 
    in our experiments which is 0.1s or 0.2s), 
    \vspace{-5pt}
    \item $N$ (\textit{write} parameter) represents the number of output tokens written at each decoding step (N=1, 2 or 3 in our experiments, output tokens can be characters or BPEs).
\end{itemize}


In details, let $\textbf{X}=(x_1, x_2,..., x_{|\textbf{X}|})$ be the source audio sequence, and $\textbf{Y}=(y_1, y_2,..., y_{|\textbf{Y}|})$ be the target hypothesis. 
Let us also assume that $g(t)$ denotes the number of  source  frames consumed by the encoder at each decoding step $t$, and $q(t)$ is the number of target tokens generated up to step $t$. In this work, $g(t)=min\{k+(t-1)*s,|\textbf{X}|\}$, and $q(t)=q(t-1)+w_t$, with $0 \leq w_t \leq N$ being the number of target tokens emitted at step $t$, and $q(0)=0$. At each decoding step $t$, the model encodes $g(t)$ source frames in order to decode at maximum N target tokens. 

Since our pre-trained offline AST models are based on Bidirectional Long Short Term Memory (BLSTM) networks, one has to re-encode from beginning of the source sequence every time new frames are read:
\vspace{-5pt}
\begin{equation}
    h^t = encode(\textbf{X}^t)
\end{equation}
with $\textbf{X}^t = x_{\leq g(t)} = (x_1, x_2, ..., x_{g(t)})$ being 
input buffer at step $t$. 

 

The decoder then takes the encoder's hidden states sequence $h^t$, and the cached previous hidden state $z_{q(t-1)}$ to compute $w_t$ hidden states, and predict $w_t$ corresponding target tokens. With $j \in [q(t-1)+1, q(t)]$, we have:


\vspace{-5pt}

\begin{equation}
    z_{j} = decode(h^t, z_{j-1}, y_{j-1})
\end{equation}


\vspace{-10pt}

\begin{equation}
    y_j = predict(z_j)
\end{equation}




We update the output buffer $\textbf{Y}^t$ by simply appending $(y_{q(t-1)+1}, ..., y_{q(t)})$ to $\textbf{Y}^{t-1}$:

\vspace{-5pt}
\begin{equation}
    \textbf{Y}^t = \textbf{Y}^{t-1} + (y_{q(t-1)+1}, ..., y_{q(t)})
\end{equation}

This process ends when the end of sequence token </s> is predicted, 
or the length of the output buffer $|\textbf{Y}|$ exceeds a threshold.\footnote{in this work, we set a $max\_length\_ratio = \frac{max\_output\_sequence\_length}{ encoder\_hidden\_state\_sequence\_length} =1.0$ for EN-DE experiments, and $1.6$ for EN-PT experiments.} In case 
the decoder generates the </s> token before the whole source sequence $\textbf{X}$ is read, we only append the tokens preceding </s>, then break the loop and read more source frames.

\begin{table}[h]
 \centering
 \caption{(BLEU / AL) scores of the EN-DE char  model evaluated on MuST-C tst-HE and MuST-C tst-COMMON. Sorted by AL of tst-HE in increasing order. AL is in $milliseconds$.}
 \begin{tabular}{ | c | c | c | c | c | c | c } 
  \hline
  \textbf{No.} & \textbf{k} & \textbf{s} & \textbf{N} & \textbf{tst-HE} & \textbf{tst-COMMON} \\ 
  \hline
  \hline
    1   &   100	&	10	&	3	&	3.01 / 743	&	4.42 / 800	\\
    \hline
    2	&	100	&	10	&	2	&	4.26	/	1049	&	6.89	/	1135	\\
    \hline
    3	&	100	&	20	&	3	&	5.49	/	1353	&	8.61	/	1441	\\
    \hline
    4	&	200	&	10	&	3	&	7.07	/	1521	&	10.37	/	1552	\\
    \hline
    5	&	200	&	10	&	2	&	8.77	/	1836	&	12.79	/	1931	\\
    \hline
    6	&	100	&	20	&	2	&	8.77	/	2062	&	12.68	/	2097	\\
    \hline
    7	&	100	&	10	&	1	&	9.46	/	2146	&	12.85	/	2157	\\
    \hline
    8	&	200	&	20	&	3	&	10.38	/	2223	&	14.6	/	2286	\\
    \hline
    9	&	200	&	20	&	2	&	13.8	/	2934	&	16.59	/	2840	\\
    \hline
    10	&	200	&	10	&	1	&	14.11	/	2973	&	16.83	/	2880	\\
    \hline
    11	&	100	&	20	&	1	&	14.64	/	3487	&	15.79	/	3086	\\
    \hline
    12	&	200	&	20	&	1	&	17.15	/	4066	&	17.94	/	3610	\\
    \hline
    \hline
    13 & \multicolumn{3}{c|}{offline} & 20.54 / 7005 &	21.38 / 5782 \\
  \hline
 \end{tabular}
\label{table:charende}
\end{table}

\vspace{-20pt}
\subsection{Evaluation of simultaneous speech translation }
\vspace{-5pt}
In order to measure the latency, we use an adaptive version \cite{simuleval2020} of Average Lagging (AL) introduced by \cite{Ma19acl}. The  original AL metric measures the average rate (in \textit{milliseconds} in our case) that the MT system lags behind an ideal wait-0
translator:
\vspace{-5pt}
\begin{equation} \label{eq:AL}
    AL_g(\textbf{X}, \textbf{Y}) = \frac{1}{\tau_g(|\textbf{X}|)} \sum_{t=1}^{\tau_g(|\textbf{X}|)}g(t)-\frac{t-1}{\gamma}
\end{equation}

The "cut-off" step $\tau_g(|\textbf{X}|) = min\{t | g(t) = |\textbf{X}|\}$ is defined as the decoding step when the policy first finishes reading all source frames, and the target-to-source length ratio $\gamma = |\textbf{Y}|/|\textbf{X}|$ is a scale factor accounting for the source and target having different sequence lengths. In our case, since the decoder can emit more than one target token at step $t$, equation (\ref{eq:AL}) becomes:
\vspace{-5pt}
\begin{equation}
    AL_g(\textbf{X}, \textbf{Y}) = \frac{1}{\tau_g(|\textbf{X}|)} \sum_{t=1}^{\tau_g(|\textbf{X}|)}[g(t)-\frac{t-1}{\gamma}]*w_t
\end{equation}

However, as observed by \cite{simuleval2020}, the original metric often generates negative values when applied to speech-to-text translation. Therefore, they propose an adaptive version which computes $\gamma$ as: $\gamma=|\boldsymbol{Y^*}|/|\textbf{X}|$, with $\boldsymbol{Y^*}$ being the reference sentence. Tables and curves presented in this paper are computed using the adaptive AL version unless stated otherwise.

In this work, 
AL is computed at word-level. This means that with models that generate target token units smaller than word, for example character or BPEs, we wait until a complete word is merged, before committing the corresponding delay to the AL computing module.\footnote{We re-use the latency class from https://github.com/facebookresearch/SimulEval for AL computing.}



\begin{table}[h]
 \centering
 \caption{(BLEU / AL) scores of the EN-PT char model evaluated on MuST-C tst-COMMON. Sorted by AL in increasing order. AL is in $milliseconds$.}
 \begin{tabular}{ | c | c | c | c | c | c } 
  \hline
  \textbf{No.} & \textbf{k} & \textbf{s} & \textbf{N} & \textbf{tst-COMMON} \\ 
  \hline
  \hline
    1	&	100	&	10	&	3	&	4.67	/	611	\\
    \hline
    2	&	100	&	10	&	2	&	8.38	/	907	\\
    \hline
    3	&	100	&	20	&	3	&	11.43	/	1237	\\
    \hline
    4	&	200	&	10	&	3	&	11.61	/	1402	\\
    \hline
    5	&	200	&	10	&	2	&	15.97	/	1748	\\
    \hline
    6	&	100	&	20	&	2	&	16.67	/	1948	\\
    \hline
    7	&	100	&	10	&	1	&	16.95	/	1976	\\
    \hline
    8	&	200	&	20	&	3	&	17.94	/	2106	\\
    \hline
    9	&	200	&	20	&	2	&	20.69	/	2697	\\
    \hline
    10	&	200	&	10	&	1	&	20.98	/	2735	\\
    \hline
    11	&	100	&	20	&	1	&	20.64	/	2910	\\
    \hline
    12	&	200	&	20	&	1	&	22.67	/	3505	\\
    \hline
    \hline
    13 & \multicolumn{3}{c|}{offline} & 25.07 / 5986 \\
  \hline
 \end{tabular}
\label{table:charenpt}
\end{table}

\vspace{-20pt}
\section{EXPERIMENTAL SETUP}
\label{sec:exp}
\vspace{-5pt}


\textbf{EN-PT pair.}
As for the EN-PT offline models, we re-use our best offline character-based (char) model, and two BPE-based (BPE400, and BPE2K) models, which were trained for IWSLT 2019 \cite{nguyen2019ontrac} shared task. As mentioned in \cite{nguyen2019ontrac}, these models were trained on a merged version of MuST-C EN-PT \cite{mustc19}, and How2 corpus \cite{sanabria18how2}. Being trained on this combination of about $674.4$ hours of training data, the char model scored $26.91$ BLEU, while the BPE400 and BPE2K model scored $24.73$, and $23.11$ BLEU respectively on MuST-C tst-COMMON set, in beam search mode ($beam\_size=10$).\footnote{Our performance in simultaneous mode will be, in contrast, given with greedy decoding mode.}

\textbf{EN-DE pair.}
The offline char EN-DE was trained  for our participation to IWSLT 2020 \cite{elbayad:hal-02895893}. MuST-C EN-DE, Europarl EN-DE \cite{europarlst}, and How2 synthetic (i.e. English speech - German synthetic translation from the original transcription) were merged together in order to train this model. It scored $23.55$ and $22.35$ BLEU on MuST-C tst-COMMON, and MuST-C tst-HE, in beam search mode ($beam\_size=10$), respectively.

All of these models were trained using normalized (mean and variance normalization) 83-dimensional Mel filter-bank and pitch features. We consistently use speed perturbation with factors of 0.9, 1.0, and 1.1 as one of the data augmentation methods in all of our experiments. Besides, \textit{SpecAugment} \cite{park2019specaugment} is used to train our EN-DE char model as well. Further details can be found in \cite{nguyen2019ontrac, elbayad:hal-02895893}.
 

\vspace{-8pt}

\section{SIMULTANEOUS DECODING RESULTS}
\label{sec:results}
\vspace{-5pt}
\subsection{Impact of decoding parameters}

The experimental results (BLEU for different AL) are given in Table \ref{table:charende} for EN-DE and in Table \ref{table:charenpt} for EN-PT for the character-based models. For a fair comparison with the online mode, the offline models were re-decoded in greedy decoding mode. This gives the results on the last rows of both the tables. We observe that: (a) the 3 parameters of our simultaneous decoding strategy allow to move over the whole range of AL (from very low latency regimes <1s to higher AL values between 2s and 3s), (b) for a latency of 2s, the strategy reaches decent BLEU scores (BLEU=
14.60 for EN-DE and BLEU=
17.94 for EN-PT on tst-COMMON), (c) writing two characters at each decoding step (N=2) with bigger stride (s=20) seems to be slightly better in term of AL, yet slightly worse in term of BLEU scores than writing one character at a time (N=1) with smaller stride (s=10) (compare lines 6-7 and 9-10 of Table \ref{table:charende} and \ref{table:charenpt} for instance).


\subsection{Impact of target granularity}



\begin{figure}
    \centering
    \begin{tikzpicture}
\pgfkeys{/pgf/number format/.cd,1000 sep={}}
\pgfmathsetmacro{\ALWue}{5806}
\pgfmathsetmacro{\Wue}{23.972}


\begin{axis}[
    height=6.0cm, width=8.0cm, 
    grid=both, y axis line style=-,
    legend style={
        font=\small, 
        legend cell align=left},
    xtick={0,500,...,4000},
    minor x tick num=1, 
    ytick={6,8,...,30},
    minor y tick num=1,
    tick label style={font=\small},
    label style={font=\small},
    xmin=0,xmax=4000,
    ymin=0, ymax=26,
    xlabel=Average Lagging (AL) in ms,
    ylabel=BLEU,
    every axis plot/.append style={line width=0.9pt, mark size=2.5pt, mark=square},
    legend to name=s2t
   ]

\foreach \x in {1000, 2000, 3000}
    \addplot [mark=none, line width=0.5pt, red, forget plot] coordinates {(\x, 0) (\x, 26)};

\addplot [mark=none, black, dashed] coordinates {(0, 25.07) (4000, 25.07)};
\addlegendentry{offline char}
\addplot [mark=none, orange, dashed] coordinates {(0, 22.7) (4000, 22.7)};
\addlegendentry{offline bpe400}
\addplot [mark=none, blue, dashed] coordinates {(0, 20.75) (4000, 20.75)};
\addlegendentry{offline bpe2k}

\addplot[black]
table [y=BLEU,x=AL]{results/blstm_char_en_pt_tst_common_maha_wait_adaptive.dat};
\addlegendentry{$char$}
\addplot[orange]
table [y=BLEU,x=AL]{results/blstm_bpe400_en_pt_tst_common_maha_wait_adaptive.dat};
\addlegendentry{$bpe400$}
\addplot[blue]
table [y=BLEU,x=AL]{results/blstm_bpe2k_en_pt_tst_common_maha_wait_adaptive.dat};
\addlegendentry{$bpe2K$}



\end{axis}

\node[anchor=south east, scale=.8] at (rel axis cs: 0.96,-0.0) {\pgfplotslegendfromname{s2t}};

\end{tikzpicture}
    \caption{Character-based vs bpe-based models on English-to-Portuguese translation, evaluated on MuST-C tst-COMMON.}
    \label{fig:charvsbpeenpt}
\end{figure}
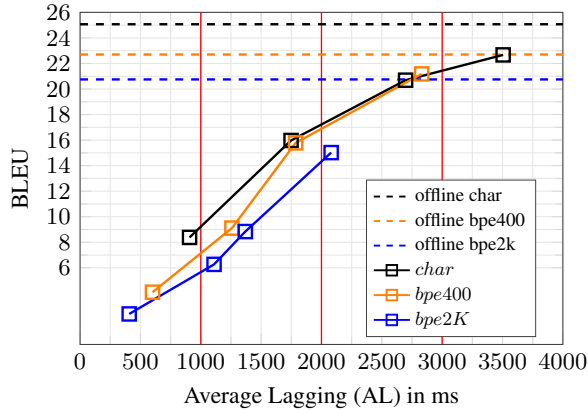

In order to investigate the impact of different target token types, we decode our pre-trained offline BPE2K and BPE400 models trained for IWSLT 2019 \cite{nguyen2019ontrac}, using the presented simultaneous decoding strategy. We alternate between different $(k, s, N)$ triplets, with $k=[100, 200]$, $s=[10, 20]$, and $N=[1, 2]$. We then sort the results by the increasing order of AL of each model, and pick the $(k, s, N)$ combinations that give the best BLEU-AL trade-offs. The results are shown in figure \ref{fig:charvsbpeenpt}, whose data points correspond to $(k, s, N) =$ $(100,10,2), (200,10,2), (200,20,2),(200,20,1)$. 
Results are given as a trend and should be taken with caution since offline char-based and BPE-based models have different performance as well.
However, figure shows that in general, for a same $(k, s, N)$ triplet, char model tends to give higher BLEU score, yet bigger AL than the two BPE models. We can observe the same trend when comparing BPE400 with BPE2K model. This could be explained by the fact that since BPEs are bigger token units than characters, when given the same amount of context (number of source frames), and forced to generate approximately the same number of target tokens, BPE models unsurprisingly give worse translation quality. However, it takes less number of source frames for BPE models to make up a word in comparison with char model, consequently resulting in smaller AL for BPE models with similar $(k, s, N)$ settings.


\subsection{Comparison to the state-of-the-art}
\label{sec:sota}

At IWSLT 2020, the winning system for speech-to-text online translation  paired an ASR system with an online MT system and decoded following a deterministic algorithm described in \cite{elbayad:hal-02895893}. The ASR system was a strong hybrid HMM/DNN system built
using the Kaldi speech recognition toolkit~\cite{Povey11thekaldi} (WER=14.2\% on MuST-C tst-COMMON dataset in offline mode). The online MT system was a Transformer-based \cite{Vaswani17nips} \textit{wait-k} decoder with unidirectional encoder. Instead of optimizing a single decoding path, \cite{elbayad:hal-02895893} jointly optimized across multiple wait-$k$ paths. 
The additional loss terms provide a richer training signal, and  yield models that  perform well under different lagging constraints. 


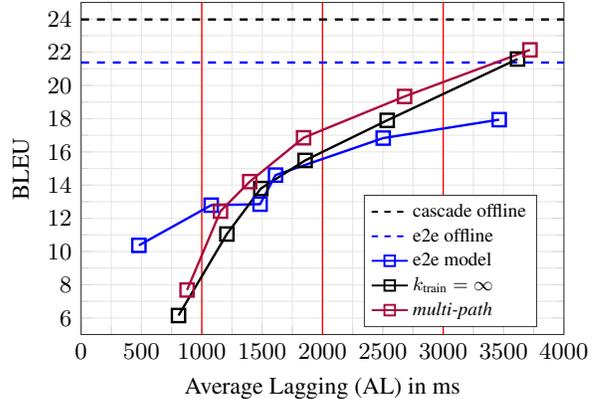
\begin{figure}
    \centering
    \begin{tikzpicture}
\pgfkeys{/pgf/number format/.cd,1000 sep={}}
\pgfmathsetmacro{\ALWue}{5806}
\pgfmathsetmacro{\Wue}{23.972}


\definecolor{p1blue}{RGB}{0, 90, 200}      
\definecolor{p1red}{RGB}{170, 10, 60}      
\definecolor{p1green}{RGB}{10, 155, 75}    
\definecolor{p1yellow}{RGB}{234, 214, 68}  
\definecolor{p1orange}{RGB}{255, 130, 95}  
\definecolor{p1purple}{RGB}{130, 20, 160}  
\definecolor{p1azure}{RGB}{0, 160, 250}    
\definecolor{padgray}{rgb}{.7,.7,.7}      
\definecolor{dimgray}{rgb}{.35,.35,.35}   
\definecolor{darkgray}{rgb}{.20,.20,.20}  

\begin{axis}[
    height=6.0cm, width=8.0cm, 
    grid=both, y axis line style=-,
    legend style={
        font=\small, 
        legend cell align=left},
    xtick={0,500,...,4000},
    minor x tick num=1, 
    ytick={6,8,...,30},
    minor y tick num=1,
    tick label style={font=\small},
    label style={font=\small},
    xmin=0,xmax=40 00,
    ymin=5, ymax=25,
    xlabel=Average Lagging (AL) in ms,
    ylabel=BLEU,
    every axis plot/.append style={line width=0.9pt, mark size=2.5pt, mark=square},
    legend to name=s2t
   ]

\foreach \x in {1000, 2000,3000}
    \addplot [mark=none, line width=0.5pt, red, forget plot] coordinates {(\x, 5) (\x, 26)};

\addplot [mark=none, black, dashed] coordinates {(0, \Wue) (4000, \Wue)};
\addlegendentry{cascade offline}

\addplot [mark=none, blue, dashed] coordinates {(0, 21.38) (4000, 21.38)};
\addlegendentry{e2e offline}

\addplot[blue]
table [y=BLEU,x=AL]{results/blstm_char_en_de_tst_common_maha_wait_adaptive.dat};
\addlegendentry{e2e model}
\addplot[black]
table [y=BLEU,x=AL]{results/tst_common_s2t_wue.dat};
\addlegendentry{$\ktr=\infty$}




\addplot[p1red]
table [y=BLEU,x=AL]{results/tst_common_s2t_multi.dat};
\addlegendentry{\emph{multi-path}}



\end{axis}

\node[anchor=south east, scale=.8] at (rel axis cs: 0.96,-0.0) {\pgfplotslegendfromname{s2t}};

\end{tikzpicture}
    \caption{Comparison of our proposed character-based EN-DE end-to-end model (\textit{e2e model}) with that of the (winning) ON-TRAC cascaded models with (\textit{multi-path}) or without (\textit{$k_{train}=\infty$}) re-training for simultaneous mode. Original AL metric \cite{Ma19acl} is used to compute the latency illustrated in this curve.}
    \label{fig:sotacompare}
\end{figure}

In Figure \ref{fig:sotacompare}, we choose the $(k, s, N)$ triplets for the EN-DE char model that give close AL regimes to that of the above cascaded models. It is clear that our proposed decoding strategy (the blue curve) performs reasonably well in comparison with the cascaded models with (\textit{multi-path}) or without (\textit{$k_{train}=\infty$}) re-training for simultaneous mode (\textit{e2e} is better in low latency regimes and \textit{cascade} is better in higher latency regimes). We recall that our \textit{e2e model} did not require re-training in  simultaneous  mode, it is thus fair to compare it with the corresponding cascaded model: $k_{train}=\infty$. We note that since \cite{elbayad:hal-02895893} use the original AL metric to compute their latency, in this curve, we also use original AL \cite{Ma19acl} to compute the latency of our end-to-end model.









\vspace{-10pt}
\section{CONCLUSION}
\vspace{-5pt}
We propose in this paper a simple yet efficient simultaneous decoding strategy, which allows pre-trained end-to-end AST offline model to decode in simultaneous mode. Our empirical evaluation, conducted on  models trained for 2  different  language  pairs  EN-DE and EN-PT, with either characters or BPEs, shows that this strategy can give decent BLEU-AL trade-offs, and its best settings  achieve competitive results in comparison with a strong cascade baseline, without re-training the models in online mode (this will however be investigated in future work).

\section{ACKNOWLEDGEMENTS}
This work was funded by the French Research Agency (ANR) through the ON-TRAC project under contract number ANR-18-CE23-0021, and was performed using HPC resources from GENCI-IDRIS (Grant 20XX-AD011011365).

\label{sec:refs}

\bibliographystyle{IEEEbib}
\bibliography{strings,refs}

\end{document}